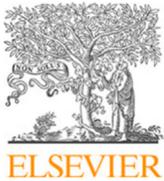

Contents lists available at ScienceDirect

# Machine Learning with Applications

journal homepage: www.elsevier.com/locate/mlwa

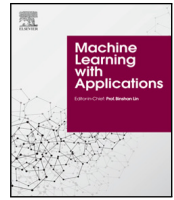

# Playing with words: Comparing the vocabulary and lexical diversity of ChatGPT and humans

Pedro Reviriego [a,*], Javier Conde [a], Elena Merino-Gómez [b], Gonzalo Martínez [c], José Alberto Hernández [c]

[a] *ETSI de Telecomunicación, Universidad Politécnica de Madrid, Spain*
[b] *Escuela de Ingenierías Industriales, Universidad de Valladolid, Spain*
[c] *Departamento de Ingeniería Telemática, Universidad Carlos III de Madrid, Spain*

## ARTICLE INFO

## ABSTRACT

The introduction of Artificial Intelligence (AI) generative language models such as GPT (Generative Pre-trained Transformer) and conversational tools such as ChatGPT has triggered a revolution that can transform how text is generated. This has many implications, for example, as AI-generated text becomes a significant fraction of the text, would this affect the language capabilities of readers and also the training of newer AI tools? Would it affect the evolution of languages? Focusing on one specific aspect of the language: words; will the use of tools such as ChatGPT increase or reduce the vocabulary used or the lexical diversity? This has implications for words, as those not included in AI-generated content will tend to be less and less popular and may eventually be lost. In this work, we perform an initial comparison of the vocabulary and lexical diversity of ChatGPT and humans when performing the same tasks. In more detail, two datasets containing the answers to different types of questions answered by ChatGPT and humans, and a third dataset in which ChatGPT paraphrases sentences and questions are used. The analysis shows that ChatGPT-3.5 tends to use fewer distinct words and lower diversity than humans while ChatGPT-4 has a similar lexical diversity as humans and in some cases even larger. These results are very preliminary and additional datasets and ChatGPT configurations have to be evaluated to extract more general conclusions. Therefore, further research is needed to understand how the use of ChatGPT and more broadly generative AI tools will affect the vocabulary and lexical diversity in different types of text and languages.

## 1. Introduction

The rapid development of tools that can generate text in a wide range of tasks, such as ChatGPT (Wu et al., 2023) is poised to change the way text is created. This has huge implications for many industries and jobs as noted by different analysts and experts (Keller, 2023; Ray, 2023). However, it may also have an impact on the future of languages. There are indeed concerns that AI tools may benefit some languages over others which may compromise the future of minority languages (Costa-jussà et al., 2022; Reviriego & Merino-Gómez, 2022). As AI-generated content populates the Internet, it will likely be used to train newer versions of generative AI tools creating a feedback loop in which AI tools depend on the content generated by previous AI-tools (Martínez et al., 2024). This can degrade the quality of newer AI tools and even lead to a collapse of AI models (Dohmatob, Feng, Yang, Charton, & Kempe, 2024; Marchi, Soatto, Chaudhari, & Tabuada, 2024; Shumailov et al., 2023). Similarly, users will be exposed to AI-generated text when they are learning languages (including their mother tongue), and thus it may influence how languages are learned.

It seems reasonable to consider that the potential impact of AI-generated text will depend on how different it is from human-generated text. Evaluating these differences is far from trivial. A few metrics such as fidelity, diversity, or the Fréchet Inception Distance (FID) have been proposed focusing on generative AI image tools (Naeem, Oh, Uh, Choi, & Yoo, 2020). However, applying them (or similar concepts) to tools such as ChatGPT is not straightforward. In this work, we consider the differences between the text generated by ChatGPT and humans focusing on a specific aspect: the vocabulary used and the lexical diversity in a given task. Words are a fundamental element of languages and the number of known words varies with age, education, and other factors (Brysbaert, Stevens, Mandera, & Keuleers, 2016). The lexicon of any language understood as the comprehensive list of all its words, forms the very foundation of human language's constitution. Words serve as the medium through which the meanings of universal knowledge are conveyed, making them an indispensable heritage. For instance, words classified as obsolete, often excluded from general dictionaries (de la Mata, 2019) of the "current language", are nonetheless crucial for understanding texts from different eras. The definitive loss of a word implies losing the essential keys needed to understand the world in its wholeness. Since the 19th century, the proliferation of dictionaries that include historical lexicography has become common in many languages.[1] Throughout the 20th century, the use of corpora [aiming] to comprehensively collect the existing lexicon in each language through written and oral language samples has become widespread.[2] In the 21st century, enriching corpora with databases has become much easier and faster, thanks to the combination of digitizing materials gathered in the previous century along with the vast amount of written and audio resources available on the Internet[3]. The size of such databases today, in addition to their different formats (written, sound, and videographic) necessarily requires electronic support.

---

* Corresponding author.
*E-mail addresses:* pedro.reviriego@upm.es (P. Reviriego), javier.conde.diaz@upm.es (J. Conde), elena.merino.gomez@uva.es (E. Merino-Gómez), gonzmart@pa.uc3m.es (G. Martínez), jahgutie@it.uc3m.es (J.A. Hernández).

Currently, online dictionaries provide us with vast repositories of words that were once considered obsolete and had disappeared from the "current use" dictionaries of all languages not too long ago (Pierre & Cerquiglini, 2018). Indeed, the removal of words that have fallen into disuse and the inclusion of new lemmas have been constant processes throughout the history of dictionaries (de la Mata, 2019). At present, thanks, among other factors, to the possibilities of digital storage and the ease and speed of word searches, the trend is to conserve every word and to preserve the knowledge associated with them.

Looking beyond individual words, it is interesting to consider how words are used in a given text (Schmitt, 2010). This has been widely studied and is related to the concept of lexical diversity that tries to capture the diversity of the vocabulary used (Kyle, 2019). A large number of metrics to quantify lexical diversity have been proposed, such as the Type-Token-Ratio (TTR), the Root-Type-Token-Ratio (RTTR) or the Corrected-Type-Token-Ratio (CTTR) that are computed from the number of words (tokens) and number of distinct words (types) in the text (Hout & Vermeer, 2007). More advanced metrics that use logarithmic relationships have also been proposed to try to obtain lexical diversity values that are independent of the length of the text, like for example, the one proposed by Maas (Tweedie & Baayen, 1998). These metrics can be used to compare the lexical diversity of different texts.

In this context, it is of interest to compare the lexical diversity of the text generated by humans and ChatGPT for a given task to understand how it may affect the evolution of the vocabulary. To the best of our knowledge, this has not been explored in detail, only the distribution of the frequency of words in AI generated languages has been studied (Diamond, 2023) showing that it follows a Zipf (Zipf, 2013) distribution as is the case in human languages.

In this paper, we take a first step in this direction by comparing the vocabulary used and the lexical diversity in human and ChatGPT answers for several types of questions and when ChatGPT is used for paraphrasing. The rest of the paper is organized as follows, Section 2 summarizes the related work and the terminology used in the paper, and then Section 3 describes the datasets used in our evaluation. The analysis is presented in Section 4 and the results, the limitations of this initial study, as well as a few ideas for future work are discussed in Section 5. The paper ends with the conclusion in Section 6.

## 2. Background

The rise of Generative AI with text-to-image generators such as DALL-E (Ramesh, Dhariwal, Nichol, Chu, & Chen, 2022) or Stable Diffusion (Esser et al., 2024b), text-to-video tools such as Sora, and in particular LLMs has transformed how interactive applications are developed with LLMs becoming a new piece in the software architecture. Current research is focused on two paths: on the one hand, increasingly larger models are being trained to achieve better results (Hoffmann et al., 2024) (e.g., Meta announced versions of LLaMA 3.1 with over 400 B parameters[4]). On the other hand, researchers are exploring model compression techniques such as quantization or neural network pruning, aiming to obtain smaller models without reducing their performance while allowing them to be loaded onto general-purpose hardware (Ma et al., 2024). Other works explore model acceleration, for instance, (Gloeckle, Idrissi, Rozière, Lopez-Paz, and Synnaeve (2024) propose simultaneous prediction of multiple tokens, thereby multiplying the model's speed and reducing its consumption.

One limitation of LLMs and conversational models is that their knowledge is limited to the dataset they have been trained on and the size of the window they can process. Some models such as Gemini 1.5 Pro have managed to increase the size of the context window by up to 10 M tokens, capable of processing around 7M words, 10.5 h of video, and 107 h of audio (Gemini Team, 2024). Training a model is very costly both computationally and economically, making it unfeasible in most cases to train a model from scratch to expand its knowledge base. However, techniques such as finetuning allow to retraining a new model starting from a base model to modify its behavior, for example, by adding more information, specializing it in a task, forcing it to respond in a specific format, or specializing it in a language. As a result, a finetune model modifies the weights of the neural network. Finetuning is cheaper than training a model from scratch, and it allows to modify the behavior of the base model, but for the use case of increasing its knowledge, it requires new training each time more information needs to be added. Techniques such as LoRA and QLoRA allow to reduce the finetuning costs by decreasing the number of parameters to be adjusted (LoRA) (Hu, Shen, Wallis, Allen-Zhu, Li, Wang, Wang, & Chen, 2021) and combining this with quantization strategies (QLoRA) (Dettmers, Pagnoni, Holtzman, & Zettlemoyer, 2023). Additionally, the new model is prone to hallucinations and is not limited to the information used for the finetuning but includes all the information with which the base model was trained. For example, if finetuning instructs a model that the capital of France is Marseille, it may still answer that the capital is Paris since the base model was trained with many datasets indicating that Paris is the capital of France. This lack of control over the response and its inability to forget the information it was trained with make it very difficult to apply an LLM for different scenarios. An alternative is Retrieval Augmented Generation (RAG) systems, in which the LLM limits its responses to the information provided at the prompt (Lewis et al., 2020). RAGs reduce hallucinations and allow control over the information the LLM will use to respond. In the case of RAGs, the text comprehension and response generation capabilities of LLMs are used. The most typical use cases for RAGs involve scenarios where information is stored in a vectorized database. However, there are other alternatives such as querying a different type of database or connecting the LLM to the internet.

---

[1] For example in Spanish https://www.rae.es/banco-de-datos/corde.

[2] For example for Spanish CREA: https://www.rae.es/banco-de-datos/crea; for English the British National Corpus (BNC): http://www.natcorp.ox.ac.uk/corpus.

[3] For English, the Oxford English Corpus https://www.sketchengine.eu/oxford-english-corpus/; for American English, the Corpus of Contemporary American English (COCA): https://www.english-corpora.org/coca/; for oral French, le corpus de la parole. In France work continues on the CRFC: corpus de référence du français contemporain (Siepmann, Bürgel, & Sascha, 2016); for Spanish on the CORPES XXI: https://www.rae.es/banco-de-datos/corpes-xxi

[4] Llama 3 400 B announcement: https://github.com/meta-llama/llama-models/blob/main/models/llama3_1/MODEL_CARD.md

One of the biggest barriers in LLM research is their evaluation due to their lack of explainability. So far, the main evaluations are based on benchmarks composed of thousands of multiple-choice questions like the Massive Multitask Language Understanding (MMLU) benchmark (Hendrycks et al., 2021). The drawback of these methods is that they only assess the model's knowledge and not its generation and language capabilities. Additionally, since these are open datasets, they can be integrated into the model's training phase, altering the evaluation results (McIntosh, Susnjak, Liu, Watters, & Halgamuge, 2024). Moreover, the computational cost of using an LLM and the quality of responses vary depending on the language. Most LLMs have been trained with datasets predominantly in English and their tokenizers and embedding models are designed to be more efficient in this language. As a result, there are studies that analyze the differences between languages in terms of evaluating their response quality (Zhang, Li, Hauer, Shi, & Kondrak, 2023), knowledge (Conde, González et al., 2024), and their cost (Ahia et al., 2023).

Advancements in generative AI extend beyond text or instructional models to other modalities. Recently, Stability AI introduced Stable Diffusion 3,[5] an image generation model based on a Multimodal Diffusion Transformer architecture, which separates the text and image modalities in independent [transformers] but integrates them during the attention phase, achieving better results than previous versions (Esser et al., 2024a). Similarly, OpenAI and Kling AI have presented the video models Sora[6] and Kling[7] that represent an advance in the state of the art of video models. They are models based on diffusion transformers capable of generating one-minute videos that maintain context across frames (Liu et al., 2024).

Over the centuries, our languages and vocabularies have evolved due to social interaction, cultural changes, historical events, and other factors (Campbell, 2013). These changes have accelerated with the advent of the Internet and other new technologies, allowing the rapid spread of new words, phrases, and idioms, breaking down linguistic barriers, and facilitating cultural exchange (Crystal, 2011; Herring, 2005; Tagliamonte & Denis, 2008). In recent years, the emergence and popularization of Generative AI, such as transformers (Vaswani et al., 2023) and ChatGPT, have introduced new ways to interact and generate text more powerfully and easily than ever before, demonstrating that these changes could accelerate even further.

The impact of generative AI on writing is already being observed in academic publications and is introducing changes in the frequencies of words (Geng & Trotta, 2024; Liang et al., 2024).[8] There are also works that use AI models to create dictionaries (Ortega-Martín et al., 2024) and more and more applications will appear in the coming years. Several works have studied the impact of conversational LLMs on the language, comparing it to humans in terms of syntax, meaning, word length, or discourse (Cai, Haslett, Duan, Wang, & Pickering, 2023). For example, Toro (2023) considers the potential phonological biases of ChatGPT while Muñoz-Ortiz, Gómez-Rodríguez, and Vilares (2023) compares several linguistic features of the text produced by humans and LLaMa (Touvron et al., 2023). The capability of LLMs to manage lexical ambiguity has also been recently studied considering both polysemy and homonymy with promising results (Ortega-Martín et al., 2023).

Other works compare the lexical diversity between ChatGPT and humans, but their main limitation is the size of the datasets, which does not allow to draw significant conclusions. The research conducted by Herbold, Hautli-Janisz, Heuer, Kikteva, and Trautsch (2023) analyzes different features like lexical diversity, however, their results are limited by their dataset which only includes 90 essays. Similarly, the work of Skowron and Baczkowska (Skowron & Bączkowska, 2023) obtains the lexical diversity of texts generated through the rephrasing of a single paper, which is not enough data for drawing relevant conclusions. To the best of our knowledge, there is no specific study of the lexical diversity of ChatGPT generated text on a dataset of relevant size and covering different types of questions and topics.

The lexical aspects of language use are complex, ideally we would like to understand both the number of words or vocabulary size, but also the depth, indicating how thoroughly those words are understood. One classical metric is the lexical density, given by the number of content words in the text, which relates to the linguistic complexity of the text. Another, more complex concept is lexical diversity which has several aspects: variability, volume, evenness, rarity, dispersion, and disparity (Jarvis & Daller, 2013). The level of sophistication of the vocabulary used is also of interest. In fact, the features, metrics, and terminology to study the lexical aspects of text is an active area of research (Johansson, 2008).

In this paper, we focus on the classical concept of lexical diversity, understood as the relation between the total number of words (tokens) and the number of distinct words (types). Other lexical features are not considered in this paper and are left for future works. The metrics used in the evaluation are the Root-Type-Token-Ratio (RTTR) (Hout & Vermeer, 2007) and the Maas metric (Tweedie & Baayen, 1998). These metrics are used to compare the lexical diversity of different texts which is the main goal of the paper.

## 3. Datasets

To compare the vocabulary used by ChatGPT and humans, we need text generated by both humans and ChatGPT for the same tasks. Interestingly, this is precisely what is done in the datasets used to evaluate algorithms to detect if a given text was generated by ChatGPT (Guo et al., 2023; Shijaku & Canhasi, 2023; Yu, Chen, Feng, & Xia, 2023). In our evaluation, we select subsets from three datasets[9] to compare the lexical diversity of texts generated by humans and by ChatGPT.

The first dataset is described in Shijaku and Canhasi (2023) contains 126 TOEFL essays written by ChatGPT and humans.[10] The essays are written in response to a question, for example: "How do movies or television influence people's behavior? Use reasons and specific examples to support your answer." The questions cover different topics and the essays have approximately 300–400 words.

The second dataset was presented in Guo et al. (2023) and contains human and ChatGPT answers to questions on different topics.[11] In our analysis, we consider four topics: Computer Science, Medicine, Finance and open questions, they have 842, 1248, 3933, and 1187 questions respectively. The questions and human answers were taken from different sources in each category, which means that they can be considered independent datasets. The answers are on average shorter, most in the range of 100–200 words.

In summary, we have five groups of 126, 842, 1248, 3933, and 1187 questions and answers to compare human and ChatGPT answers. Note that different from other papers, we are not interested in discriminating if a given answer was generated by ChatGPT or a human but in comparing the vocabulary used by both when doing the same tasks. To that end, rather than individual responses to questions, the entire corpus of responses to each type of question is more useful. Therefore, we have merged all the answers for a given topic in a single file for

---

**Table 1**
Summary of the data used in the evaluation.

| Category | Type | Elements | Word per element |
|---|---|---|---|
| TOEFL | Answers | 126 | 200–400 |
| Medicine | Answers | 1248 | 100–200 |
| Computing | Answers | 842 | 100–200 |
| Finance | Answers | 3933 | 100–200 |
| Open questions | Answers | 1187 | 100–200 |
| Questions | Paraphrase | 172959 | 10–30 |
| Sentences | Paraphrase | 247138 | 10–30 |

ChatGPT and in another for humans; those files are used for vocabulary analysis in the next section.

In addition to these two datasets that contain responses to questions, a third dataset (Vladimir Vorobev, 2023) with paraphrases of short questions and sentences has also been evaluated. In this case, ChatGPT is asked to generate five paraphrases for each input sentence or question. The inputs are much shorter, most having fewer than 20 words. There are two categories: sentences and questions with 172,059 and 247,138 elements, respectively. For each category, all human elements are merged in a single file. The same is done for ChatGPT but including only one of the paraphrases to have a fair comparison as including several paraphrases of the same sentence intuitively reduces the number of distinct words compared to having unrelated sentences. The data used in the evaluation are summarized in Table 1, all the datasets used have been generated using ChatGPT-3.5 with its default configuration parameters.

## 4. Vocabulary analysis

Once all the answers for a given category are merged into a single file, a standard preprocessing through the python's Natural Language Toolkit[12] (NLTK) is applied.[13] First, we tokenized the text, that is, we divided the text into words. Then, we carried out a Part-Of-Speech (POS) tagging (Chai, 2022) process by assigning the respective grammatical category to each token considering the text context. For the experiment, we used the Penn Treebank Tagset (Marcus, Santorini, & Marcinkiewicz, 1993) and the Universal Tagset (Petrov, Das, & McDonald, 2012) versions of NLTK. For example, the sentence ''He met six friends from Spain and seven from Germany.'' is tokenized and POS tagged into the following structure (Word, Penn grammatical category, Universal grammatical category):

```
("He", "PRON", "PRP"),
("met", "VERB", "VBD"),
("six", "NUM", "CD"),
("friends", "NOUN", "NNS"),
("from", "ADP", "IN"),
("US", "NOUN", "NNP"),
("and", "CONJ", "CC"),
("seven", "NUM", "CD"),
("from", "ADP", "IN"),
("Germany", "NOUN", "NNP")
(".", ".", ".")
```

Then, we lemmatized the tagged tokens using the English lexical database WordNet (Miller, 1995) (included in NLKT) and we removed all punctuation. In the example shown, the result would be the following:

```
("He", "PRON", "PRP"),
("meet", "VERB", "VBD"),
("six", "NUM", "CD"),
("friend", "NOUN", "NNS"),
("from", "ADP", "IN"),
("US", "NOUN", "NNP"),
("and", "CONJ", "CC"),
("seven", "NUM", "CD"),
("from", "ADP", "IN"),
("Germany", "NOUN", "NNP")
```

After that, the total number of words ($n$) and of distinct words ($t$) and their frequencies have been computed. Classical lexical diversity metrics use both the number of words ($n$) and of distinct words ($t$). For example, the Type-Token-Ratio (TTR) computes the fraction of distinct words ($t$) over the total words ($n$). However, this results in a metric that depends on the length of the text as the number of distinct words tends to grow less than the number of words with text length. Over the years, more complex metrics have been proposed to compensate for this effect for example by using the square root of the number of total words or by using more complicated formulas. In our evaluation, two such metrics, the Root-Type-Token-Ratio (RTTR) and Maas have been used (Hout & Vermeer, 2007).

The RTTR is computed as follows:

$$RTTR = \frac{t}{\sqrt{n}} \tag{1}$$

and grows with lexical diversity. Instead, the Maas metric is given by:

$$Maas = \frac{log(n) - log(t)}{(log(n))^2} \tag{2}$$

decreases with lexical diversity. Both are less dependent on text length than the classical TTR metric.

The same metrics were applied by removing the most common English stop words (Chai, 2022). In the example, the resultant words would be:

```
("meet", "VERB", "VBD"),
("six", "NUM", "CD"),
("friend", "NOUN", "NNS"),
("US", "NOUN", "NNP"),
("seven", "NUM", "CD"),
("Germany", "NOUN", "NNP")
```

Then, we applied the same metrics by selecting only nouns, verbs, adverbs, and adjectives.

```
("meet", "VERB", "VBD"),
("friend", "NOUN", "NNS"),
("US", "NOUN", "NNP"),
("Germany", "NOUN", "NNP")
```

The RTTR and Maas results of the experiment are summarized in Tables 2, 3, and 4 for all lemmas, after removing punctuation, common stop words, and selecting only nouns, verbs, adverbs, and adjectives, respectively. The trends comparing humans and ChatGPT are similar in all cases, but the values for all three phases of the text processing are provided for completeness. It can be observed that while the number of words or distinct words is higher for humans or for ChatGPT[14] depending on the category, the RTTR and Maas metrics indicate that the lexical diversity of human answers is higher for all categories. In more detail, the RTTR is consistently higher for human answers and the Maas metric is consistently lower for human answers. This holds both when keeping and removing commonly used stop words and when selecting only nouns, verbs, adverbs, and adjectives. Depending on the category, the words include technical terms or names of places and persons, analyzing the number of distinct words by those types of words would also be of interest. Fig. 1 shows the RTTR and metrics in graphical form. It can be seen that the largest differences between human and ChatGPT texts occur when answering questions (TOEFL, Medicine, Computing, Finance, and Open Questions) while the differences are smaller for paraphrasing. This may be related to the nature of the tasks, answering questions seems to provide more freedom to use different vocabulary than paraphrasing. This suggests that the lexical diversity differences will depend on the task (instructions, questions, summarization, translation, paraphrasing, etc.).

---

[14] Note that the medicine, finance, and open question categories have several ChatGPT answers for some of the questions, see Table 1 in Guo et al. (2023).

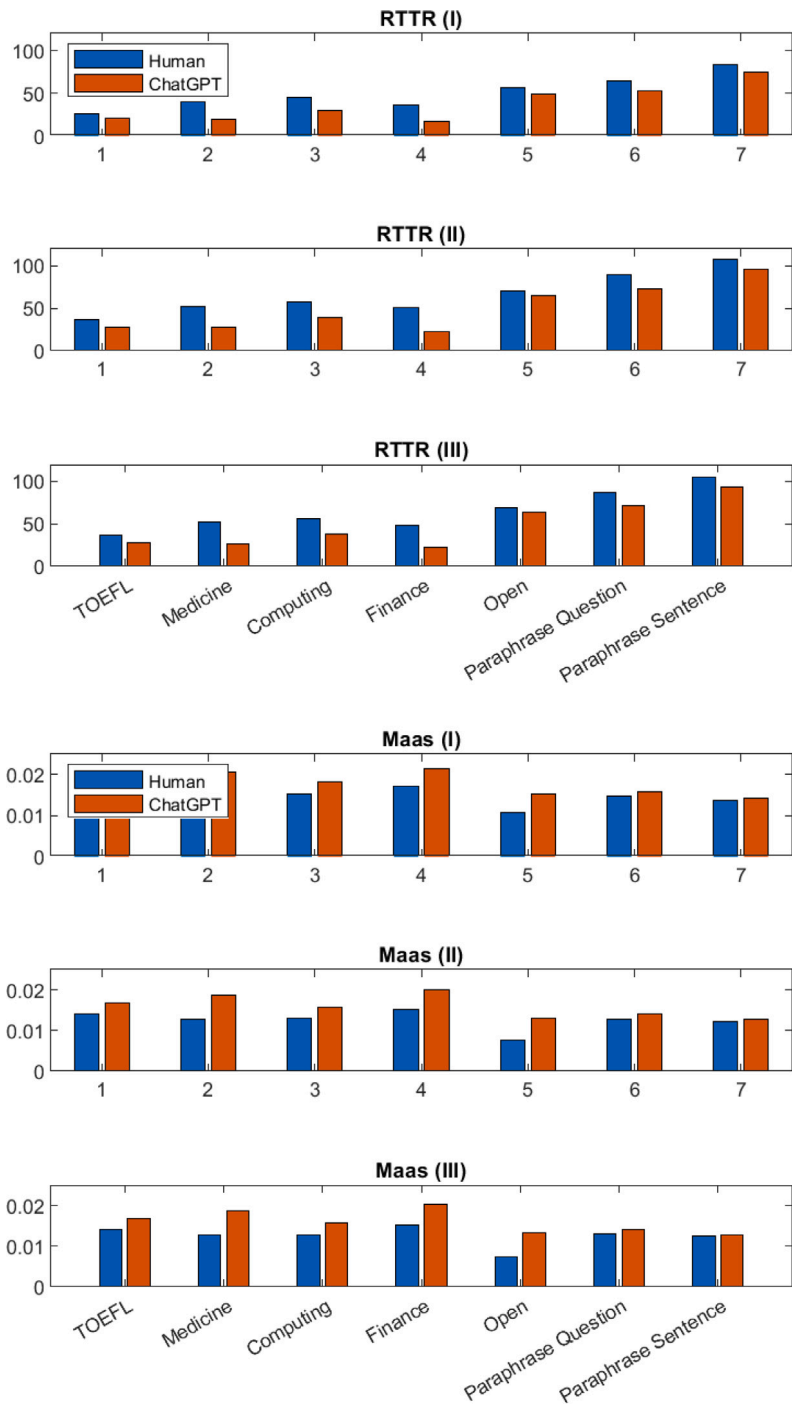

**Fig. 1.** Comparison of RTTR (top) and Maas (bottom) metrics for humans (blue) and ChatGPT (red). Larger RTTR and lower Maas correspond to higher diversity. (I): All words. (II): Without stop words. (III): Without stop words and only verbs, adverbs, and adjectives. (For interpretation of the references to color in this figure legend, the reader is referred to the web version of this article.)

Finally, although our focus is on lexical diversity, we also calculate the lexical density, that is, the proportion of nouns, verbs, adjectives, and adverbs (also called content words) present in the text (Johansson, 2008). A text with more lexical density provides more information with the same amount of words. Table 5 shows the results of the lexical density for each category. This metric does not provide significant conclusions as ChatGPT exhibits higher lexical density in some categories, while humans do so in others.

In summary, these initial results seem to suggest that the 3.5 version of ChatGPT used to generate those datasets tends to use fewer distinct words than humans and has smaller lexical diversity. However, the

**Table 2**

Number of words, number of distinct words, RTTR, and Maas metrics for each category. In bold the largest (lowest) value for each (the Maas) metric.

| Category | Type | Number of words | Distinct words | RTTR | Maas |
|---|---|---|---|---|---|
| TOEFL | Human | **49768** | **5901** | 26.45 | **0.0182** |
| TOEFL | ChatGPT | 41784 | 4229 | 20.69 | 0.0202 |
| Medicine | Human | 106180 | **13133** | 40.30 | **0.0156** |
| Medicine | ChatGPT | **250230** | 10183 | 20.36 | 0.0207 |
| Computing | Human | 167410 | **18587** | 45.43 | **0.0152** |
| Computing | ChatGPT | 155803 | 11743 | 29.75 | 0.0181 |
| Finance | Human | 707628 | **31114** | 36.99 | **0.0172** |
| Finance | ChatGPT | **944355** | 16299 | 16.77 | 0.0214 |
| Open questions | Human | 35126 | 10585 | 56.48 | **0.0109** |
| Open questions | ChatGPT | **419462** | **31755** | 49.03 | 0.0154 |
| Paraphrase of questions | Human | 2781777 | **71146** | 63.91 | **0.0148** |
| Paraphrase of questions | ChatGPT | **3105856** | 65175 | 52.51 | 0.0157 |
| Paraphrase of sentences | Human | **3771366** | **119569** | 83.83 | **0.0137** |
| Paraphrase of sentences | ChatGPT | 3577265 | 107096 | 74.69 | 0.0142 |

**Table 3**

Results after removing stop words. Number of words, number of distinct words, RTTR, and Maas metrics for each category. In bold the largest (lowest) value for each (the Maas) metric.

| Category | Type | Number of words | Distinct words | RTTR | Maas |
|---|---|---|---|---|---|
| TOEFL | Human | 24020 | **5665** | 36.55 | **0.0141** |
| TOEFL | ChatGPT | 21733 | 4081 | 27.66 | 0.0168 |
| Medicine | Human | 60319 | **12847** | 52.31 | **0.0128** |
| Medicine | ChatGPT | **134542** | 10001 | 27.27 | 0.0186 |
| Computing | Human | 103237 | **18370** | 57.17 | **0.0130** |
| Computing | ChatGPT | 86685 | 11557 | 39.25 | 0.156 |
| Finance | Human | 368636 | **30642** | 50.47 | **0.0151** |
| Finance | ChatGPT | **500566** | 16040 | 22.67 | 0.0200 |
| Open questions | Human | 22273 | 10422 | 69.83 | **0.0076** |
| Open questions | ChatGPT | **233756** | **31527** | 65.21 | 0.0131 |
| Paraphrase of questions | Human | 1423867 | **106703** | 89.42 | **0.0129** |
| Paraphrase of questions | ChatGPT | **1598740** | 92551 | 73.20 | 0.0140 |
| Paraphrase of sentences | Human | **2279774** | **163500** | 108.29 | **0.0123** |
| Paraphrase of sentences | ChatGPT | 2174323 | 141779 | 96.15 | 0.0128 |

**Table 4**

Results after removing stop words and selecting only nouns, verbs, adverbs, and adjectives. Number of words, number of distinct words, RTTR, and Maas metrics for each category. In bold the largest (lowest) value for each (the Maas) metric.

| Category | Type | Number of words | Distinct words | RTTR | Maas |
|---|---|---|---|---|---|
| TOEFL | Human | 23005 | **5549** | 36.59 | **0.0141** |
| TOEFL | ChatGPT | 21072 | 3997 | 27.53 | 0.0168 |
| Medicine | Human | 57910 | **12542** | 52.12 | **0.0127** |
| Medicine | ChatGPT | **132020** | 9757 | 26.85 | 0.0186 |
| Computing | Human | 97135 | **17633** | 56.58 | **0.0129** |
| Computing | ChatGPT | 84681 | 11218 | 38.55 | 0.0157 |
| Finance | Human | 343135 | **28530** | 48.70 | **0.0153** |
| Finance | ChatGPT | **485529** | 15306 | 21.97 | 0.0202 |
| Open questions | Human | 20362 | 9777 | 68.52 | **0.0075** |
| Open questions | ChatGPT | **221897** | **30044** | 63.78 | 0.0132 |
| Paraphrase of questions | Human | 1349778 | **101495** | 87.36 | **0.0130** |
| Paraphrase of questions | ChatGPT | **1515505** | 88264 | 71.70 | 0.0140 |
| Paraphrase of sentences | Human | **2110285** | **152653** | 105.08 | **0.0124** |
| Paraphrase of sentences | ChatGPT | 2009146 | 133187 | 93.96 | 0.0129 |

results are very preliminary, and further investigation is needed to understand how the different parameters influence the vocabulary used by humans and ChatGPT. For example, the task, the persons, the settings and version of ChatGPT, and the questions.

In the last experiment, to get an initial understanding of the impact of the ChatGPT version on lexical diversity, the first five datasets (TOEFL, Computer Science, Medicine, Finance, and open questions) were run with the latest version of ChatGPT-4[15] and the results are summarized in Fig. 2. It can be observed that the newer version improves the lexical diversity which is closer to that of humans and, in some cases, even larger depending on the dataset and metric. This suggests that as models evolve, they may have the same or even better lexical diversity than humans.

## 5. Discussion

As ChatGPT-generated content becomes increasingly prevalent on the Internet, it is of interest to examine the differences in vocabulary used by ChatGPT (and similar AI tools) compared to human-generated

---

[15] The model used was gpt-4o-2024-05-13.

**Table 5**

Results of Lexical density for each category.

| Category | Type | Content words | Total words | Lexical density |
|---|---|---|---|---|
| TOEFL | Human | **23005** | **49768** | 0.462 |
| TOEFL | ChatGPT | 21072 | 41784 | **0.504** |
| Medicine | Human | 57910 | 106180 | **0.545** |
| Medicine | ChatGPT | **132020** | **250230** | 0.528 |
| Computing | Human | **97135** | **167410** | **0.580** |
| Computing | ChatGPT | 84681 | 155803 | 0.544 |
| Finance | Human | 343135 | 707628 | 0.485 |
| Finance | ChatGPT | **485529** | **944355** | 0.514 |
| Open questions | Human | 20362 | 35126 | **0.580** |
| Open questions | ChatGPT | **221897** | **419462** | 0.529 |
| Paraphrase of questions | Human | 1349778 | 2781777 | 0.485 |
| Paraphrase of questions | ChatGPT | **1515505** | **3105856** | **0.488** |
| Paraphrase of sentences | Human | **2110285** | **3771366** | 0.560 |
| Paraphrase of sentences | ChatGPT | 2009146 | 3577265 | **0.562** |

content. Understanding these differences is crucial, as the vocabulary used by AI tools may impact language learning and usage among individuals, as well as influence the development and training of future AI models. However, this is challenging as the vocabulary used depends on the person, the tasks, the AI model version, the parameters, etc., leading to a complex scenario. Our preliminary comparison of the vocabulary used and lexical diversity of ChatGPT and humans when answering different categories of questions and when paraphrasing sentences and questions shows that ChatGPT consistently uses a smaller vocabulary and has lower lexical diversity. However, our results are based on a minimal set of data and a single AI tool and version (ChatGPT-3.5) and thus have to be taken with caution. In fact, our initial experiments with ChatGPT-4 show an increase in lexical diversity which in some cases exceeds that of humans.

Further research is needed to check if these observations are generally applicable to other tasks, tools, and languages and to find additional insights into how ChatGPT and humans use the vocabulary in different scenarios. To make this possible, further datasets containing text generated by ChatGPT and humans for the same tasks are needed. One possibility is to use datasets specifically created to evaluate algorithms that detect text written by AI tools. However, this has limitations as the datasets are not designed to test the vocabulary. Therefore, it would also be of interest to create datasets to evaluate the vocabulary of AI tools. More generally, evaluating the vocabulary of AI tools seems to be a new and interesting area for research in the intersection of computing and linguistics.

It seems that automating the analysis is needed to perform tests at scale. A possible avenue to automate the analysis could be to take human written texts, such as books, and ask ChatGPT to rewrite them and then compare the number of distinct words. The translation from one language to another and then comparing it with a human translation could also be another avenue for automation. Developing software to process entire books or even larger sets of texts using the ChatGPT API would enable the evaluation of different versions, parameters (including role), and prompts for ChatGPT to better understand how those settings impact the vocabulary used by ChatGPT and how it evolves across versions of the tool.

This work has only considered lexical diversity. As discussed in the introduction, there are other lexical features that are worth studying to understand how the text generated by AI tools compares with texts generated by humans. Evaluating the performance of AI tools in lexical-related tasks such as disambiguation is another interesting study (Ortega-Martín et al., 2023). The potential biases introduced by AI-generated text at the lexical level should also be considered and evaluated. The analysis of these and other lexical aspects is the natural continuation of this paper and is left for future work.

The key points from the results and discussion can be summarized as follows:

1. Generative AI tools that generate text such as ChatGPT will produce an important fraction of the text in the future.
2. As text generation concentrates on a few AI tools their impact on language evolution can be significant.
3. Mechanisms to identify which content has been generated by AI models, with which version, and for what purpose would help measure the impact of AI in text generation (Conde, Reviriego, et al., 2024).
4. Therefore it is important to assess how the AI generated text compares with human text, something that is neglected in most evaluation benchmarks.
5. One relevant feature is lexical diversity, a lower diversity in AI generated text could lead to an impoverishment of the language.
6. Preliminary evaluation results show that initial versions of ChatGPT have lower lexical diversity than humans, however the latest versions have improved with diversity approaching that of humans.
7. Given the importance of language, evaluation benchmarks to assess linguistic features of AI generated text should be designed and used as part of AI model evaluations.

## 6. Conclusions

In this work, we have compared the vocabulary used and lexical diversity of ChatGPT and humans when performing the same tasks. The results show that in the datasets considered, ChatGPT uses a smaller number of distinct words. However, these results are very preliminary, and more research is needed to understand the differences in the vocabulary used by AI tools such as ChatGPT and humans. Those may depend on the tasks, person, version, and configuration of the AI tool among other parameters, and thus additional datasets are needed to provide a comprehensive evaluation of the vocabulary used.

Analyzing the vocabulary used by AI tools is a new topic for research in the intersection of computing and linguistics that we think deserves further attention as it may have implications for the evolution of languages in the future. For example, as persons are exposed to AI-generated text, how will this influence the learning and use of the language? This will, among other things, depend on the vocabulary used by AI tools. This vocabulary will also influence the frequency of words and probably their future as those not used by AI tools may become less and less used.

To analyze the vocabulary of ChatGPT, software tools that automate the process of testing are needed. In more detail, tools that take large amounts of human text and interact with ChatGPT to produce the equivalent text generated by the AI model. Once those tools are available, AI models such as ChatGPT could be compared on their vocabulary for different tasks or languages as they are now compared on other performance metrics. The comparison can also be done for a wide range of lexical features to have a better understanding of the lexical capabilities of AI models.

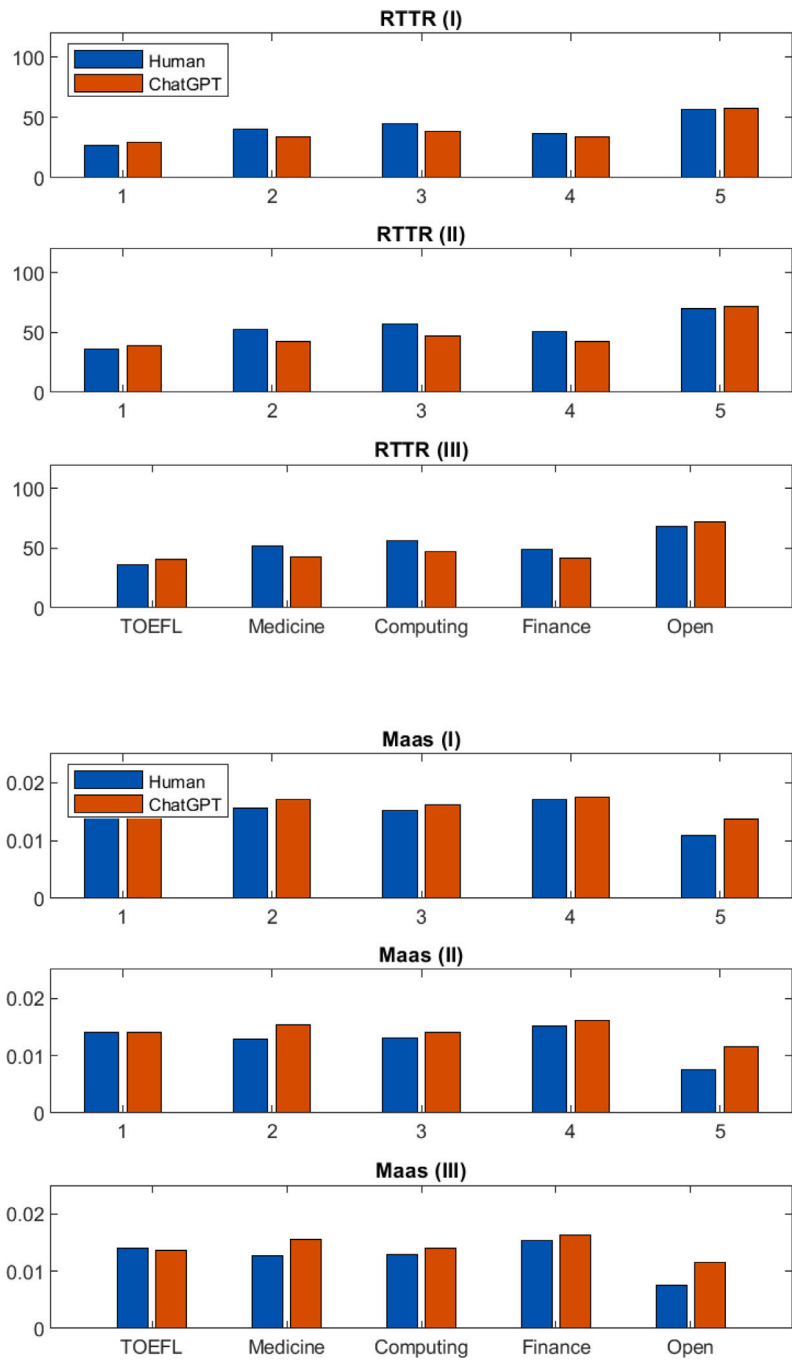

**Fig. 2.** Comparison of RTTR (top) and Maas (bottom) metrics for humans (blue) and ChatGPT-4 (red). Larger RTTR and lower Maas correspond to higher diversity. (I): All words. (II): Without stop words. (III): Without stop words and only verbs, adverbs, and adjectives. (For interpretation of the references to color in this figure legend, the reader is referred to the web version of this article.)

The main conclusion of this work is that as the amount of text generated by AI tools increases, it will have an impact on the evolution of the language. Therefore, evaluating the linguistic features of AI generated text as part of AI model evaluation is needed to ensure that AI generated text does not lead to an impoverishment of the language.

**CRediT authorship contribution statement**

**Pedro Reviriego:** Conceptualization, Methodology, Writing - Original Draft, Writing – review & editing, Supervision. **Javier Conde:** Software, Validation, Data Curation, Writing – original draft, Writing – review & editing. **Elena Merino-Gómez:** Conceptualization, Writing – original draft, Writing – review & editing. **Gonzalo Martínez:** Software, Validation, Data Curation, Writing – original draft, Writing – review & editing. **José Alberto Hernández:** Conceptualization, Methodology, Writing - Original Draft, Writing - Review & Editing, Supervision.

**Declaration of competing interest**

The authors declare that they have no known competing financial interests or personal relationships that could have appeared to influence the work reported in this paper.

**Acknowledgments**

This work was supported by the FUN4DATE (PID2022-136684OB-C22) and ENTRUDIT (TED2021-130118B-I00) projects funded by the Spanish Agencia Estatal de Investigación (AEI), Spain and by the OpenAI Research Access Program, US.

**Data availability**

No data was used for the research described in the article.